# An Adaptive Stigmergy-based System for Evaluating Technological Indicator Dynamics in the Context of Smart Specialization

Antonio L. Alfeo[1], Francesco P. Appio[2], Mario G. C. A. Cimino[1],
Alessandro Lazzeri[1], Antonella Martini[2], and Gigliola Vaglini[1]

[1]*Department of Information Engineering, Università di Pisa, largo Lazzarino 1, Pisa, Italy*
[2]*Department of Energy, System, Territory and Construction Engineering, Università di Pisa, largo Lazzarino 1, Pisa, Italy*
alfeo.luca@gmail.com, fp.appio@ing.unipi.it, mario.cimino@unipi.it,
alessandro.lazzeri@for.unipi.it, a.martini@ing.unipi.it, gigliola.vaglini@unipi.it

Abstract: Regional innovation is more and more considered an important enabler of welfare. It is no coincidence that the European Commission has started looking at regional peculiarities and dynamics, in order to focus Research and Innovation Strategies for Smart Specialization towards effective investment policies. In this context, this work aims to support policy makers in the analysis of innovation-relevant trends. We exploit a European database of the regional patent application to determine the dynamics of a set of technological innovation indicators. For this purpose, we design and develop a software system for assessing unfolding trends in such indicators. In contrast with conventional knowledge-based design, our approach is biologically-inspired and based on self-organization of information. This means that a functional structure, called track, appears and stays spontaneous at runtime when local dynamism in data occurs. A further prototyping of tracks allows a better distinction of the critical phenomena during unfolding events, with a better assessment of the progressing levels. The proposed mechanism works if structural parameters are correctly tuned for the given historical context. Determining such correct parameters is not a simple task since different indicators may have different dynamics. For this purpose, we adopt an adaptation mechanism based on differential evolution. The study includes the problem statement and its characterization in the literature, as well as the proposed solving approach, experimental setting and results.

## 1 INTRODUCTION AND MOTIVATION

After years of economic crisis and the resulting reduction of resources available for research and development investments, Smart Specialization has immediately become a very relevant concept to get these two questions answered (Foray, 2013). It represents an important chance for a progressive economical restart. In order to develop a policy-prioritization logic to foster regional growth is important to have a deep knowledge of the potential evolutionary pathways related with the existing dynamics and the structures at regional level (McCann and Raquel Ortega-Argilès, 2013). In this light, each region should start this process using as standpoints the knowledge-based sectors in which already presents a consistent 'critical mass' or, at least, capabilities that refer to a future potential exploitable with right and focused investments.

In the last decade, several causes have determined the increasing need for rationalization of resources within regions. The crucial ones are the increased globalization, mainly pursued by multinational enterprises, the economic crisis involving all EU regions with different magnitudes and the diffusion of a new wave of general purpose technologies. This situation calls for a deep rethinking of the overall approach to regional development; policy-makers and experts largely agree on the fact that the new economic boost should originate exploiting and enhancing the specific potential and competitive advantage of each region through focused innovation policies. On this line, the European Commission has established a program labelled 'Smart Specialization', consisting in a set of policies and guidelines aimed to promote the efficient and effective use of public investment in research and development (R&D).

Smart Specialization is defined as "*an industrial and innovation framework for regional economies that aims to illustrate how public policies, framework conditions, but especially R&D and innovation investment policies can influence economic, scientific and technological specialization of a region and consequently its productivity, competitiveness and economic growth path. It is a logical continuation in the process of deepening, diversifying and specializing of more general innovation strategies, taking into account regional specificities and inter-regional aspects, and thus a possible way to help advanced economies (as well as emerging economies) to restart economic growth by leveraging innovation led / knowledge-based investments in regions*" (OECD 2013, p.17). From one hand, this approach requires the concentration of R&D resources in few domains; from the other hand, a consisting part of literature underlines the importance of industry diversification in promoting innovation. In this light, the dichotomy specialization-diversification has become topical.

The long term aim of this work is exploring whether - and to what extent - different policies of 'technological specialization' and 'technological diversification' pays off in term of wealth creation at regional level. Then, we want to provide policy makers with computerized support in the analysis of innovation-relevant trends (Jin, 2014). To properly move into that direction, we start looking at this problem by analysing the trends of the aforementioned indicators for 268 EU-27 regions over 35 technological domains in the period 1990-2012, in order to obtain a model that can efficiently recognize significant events. For this purpose, we have designed and developed a software system. In contrast with conventional knowledge-based design, our approach is biologically-inspired and based on stigmergy as a mechanism of self-organization of information. Moreover, the performance of such a model is contrasted with a supervised adaptation based on the Differential Evolution (DE hereafter).

## 2 BIOLOGICALLY-INSPIRED DATA ANALYSIS

In this paper we propose to use the principles of the stigmergy for assessing unfolding trends in time variant indicators. In biology, stigmergy is an indirect communication mechanism between individuals of an insect society. In marker-based stigmergy (Parunak, 2006) volatile substances, such as pheromones, maintain the information locally for other individual to perceive. In computer science, marker-based stigmergy can be employed as a powerful computing paradigm exploiting both spatial and temporal dynamics, because it intrinsically embodies the time domain (Cimino, 2015). Moreover, marker-based stigmergy can be considered a computational black box modelling approach, because no domain model is assumed at design time and then results are not directly interpretable.

In Figure 1 we present the terminology via an ontology diagram. Concepts are enclosed in white ovals and connected by properties (represented as black directed edges). A property that cannot be directly sensed (i.e., instantiated) is represented as an abstract property, shown by a dashed edge.

More specifically, it is known that diversification and specialization of *Patents applied in a Region measure the Innovation* of the region itself. Thus, it is important for a *Policy Maker to analyse Trends of Innovation,* to properly address the investments. Such trends cannot be directly sensed nor associated to the Innovation. For this purpose, there are three important *indicators which quantify Innovation*: *specialization (S)*, *related variety (R),* and *unrelated variety (U).* The study of such *Trends* by the *Policy Maker* is fundamental to recognize scenarios of interest, i.e., the ways in which special situations may develop. Example of scenarios of interest are: (i) R or U decreases, while S increases; (ii) R or U decreases, while S is stable; (iii) R or U increases, while S is stable;(iv) R or U increases, while S increases.

The problem is to detect variations of an indicator in terms of increase, decrease or stability. In this paper, we adopt an emergent modeling perspective. With an emergent approach, the focus is on the low level processing. In Figure 1 we also present the terminology related to the approach. More specifically, an *Indicator Value enables* the release of a *Mark. Marks aggregate in Tracks*, depending on their spatiotemporal local dynamics. Emergent paradigms are based on the principle of the self-organization of the data, which means that a functional structure, the *Track*, appears and stays spontaneous at runtime when local dynamism occurs. A particular *Track* representing only the main characteristics of such local dynamics is the *Prototype*. It is the *Dissimilarity* which *compares Prototypes* generated at different times, in order to *assess the Trend*. Finally, the *Evolution* process *adapts Mark, Track, Prototype* and *Dissimilarity* to properly fit the temporal dynamics of the indicators. The Evolution process represents the application of biologically-inspired patterns to adapt parameters. The approach iteratively tries to improve a population of candidate parameters with regard to a given measure of quality, or fitness. Solutions are found by means of transformation mechanisms inspired by biology, such as reproduction, mutation, recombination, selection, in an environment where competition is represented by the quality measure.

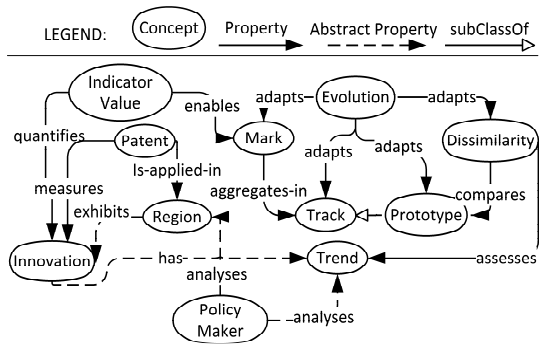

Figure 1: Ontological view of the approach.

## 3 ARCHITECTURAL DESIGN

Figure 3 illustrates the system architecture, made of four main subsystems, i.e., *marking*, *trailing*, *prototyping*, and *dissimilarity*. At the input/output interfaces of each subsystem, an *unbiasing* module is used for a better efficiency and alignment of the processing layers. This lets an input signal to reach a certain level before a processing layer passes it to the next layer, and allows a better distinction of the critical phenomena during unfolding events, with a better detection of the progressing levels.

The marking subsystem transforms input data into *marks*, whereas the trailing subsystem aggregates and evaporates marks as a *track* in the stigmergic space. The prototyping subsystem provides a simplified version of the track. It is a vehicle of abstraction, leading to the emergence of high-level information. The dissimilarity subsystem evaluates the difference between consecutive prototypes in order to extract trend information of the indicators. The proposed mechanism works if structural parameters are correctly adapted for the given application context. Determining such correct parameters is not a simple task since different indicators may have different dynamics. For this purpose, we adopt a tuning mechanism based on the DE. In the next subsections each module and subsystem is precisely described, by using a pilot data sample.

*The Unbiasing Module*
Figure 2 shows the U indicator of a region for 4 years, in dashed line. To unbias the input signal the *s-shaped function* is used, having the following behaviour: input values smaller\larger than $(\beta - \alpha)/2$ are lowered\raised; values smaller\larger than $\alpha$\$\beta$ assume the minimum\maximum value, i.e., 0\1. In biologically inspired subsystems, this function models the active zone of a signal generated by a subsystem (Avvenuti, 2013). Figure 2 shows in thick line the unbiasing output, with $\alpha_M$=0.2, $\beta_M$=0.8.

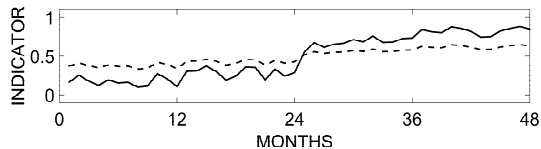

Figure 2: An example of application of unbiasing.

*The Marking Subsystem*
The Marking takes an unbiased sample $\tilde{d}(t)$ of a normalized input time series *D*, and releases a *mark* in a marking space whose codomain is called *intensity* (Cimino 2015). The mark has four structural attributes: the center position $\tilde{d}(t)$, the intensity *I*, the *mark extension* $\varepsilon$, and the *mark evaporation* $\theta$. Figure 4 (a) shows, in thick line, the mark released by the sample $\tilde{d}(2)$ of our pilot time series. The mark shape is an isosceles triangle: its center is $\tilde{d}(2)$=0.25, its height *I*=1, and its base has length $2\varepsilon$=0.5.

*The Trailing Subsystem*
The evaporation $\theta$ is the temporal decay of the mark. After each step the mark intensity decreases by a percentage $\theta$. Thus, evaporation leads towards a progressive disappearance of the mark. Anyway, subsequent marks can reinforce previous mark in the environment if their shapes overlap. In Figure 4 (b) we also show the mark $\tilde{d}(2)$ after an evaporation step, in thin line. In Figure 4 (right) we show in thin line three consecutive marks, their apex coordinates (x, y). We also show the final *track*, $T_3$ at time *t*=3, in thick line, as the sum of the track intensities

*The Prototyping Subsystem*
The Prototyping subsystem takes as input the output track of the trailing subsystem, $T_t$. This input is first unbiased, as $\tilde{T}_t$. The prototype $P_t$ is then generated as a triangular shape, with base width $2\varepsilon$, saturation height $I_{max}$=I/(1-$\theta$) (Avvenuti, 2013), and center $p_t$. Figure 5 shows in dotted line the track $T_3$ of Figure 4 (b), the unbiased track $\tilde{T}_3$ in dashed line, and the corresponding prototype $P_3$ centered in $p_3$ in solid line. The center $p_t$ of the prototype is the position that maximizes the similarity between the unbiased track and the prototype itself.

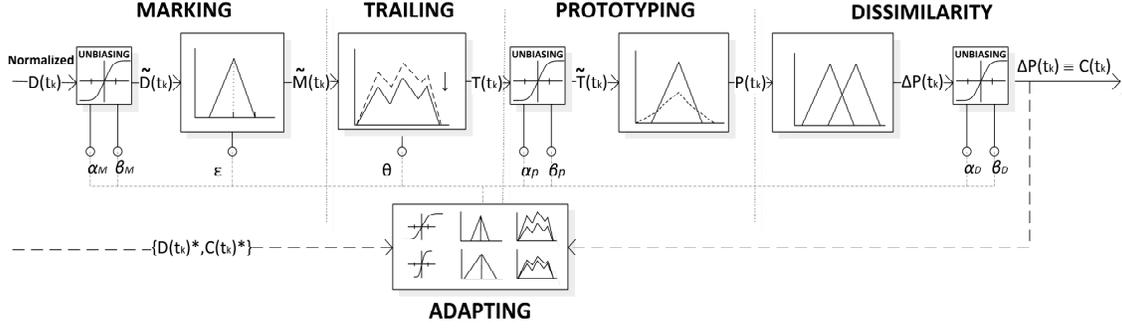

Figure 3: Architectural overview of our data analysis system based on Stigmergy.

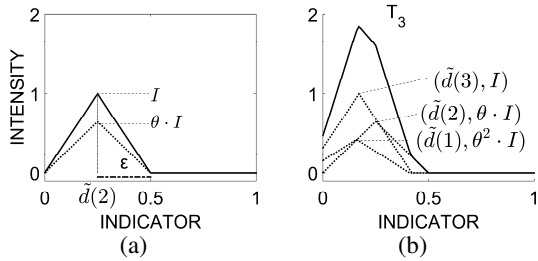

Figure 4: Single mark shape (a) and aggregation of three marks (b) in a track.

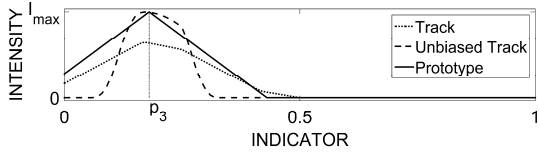

Figure 5: An example of prototyping.

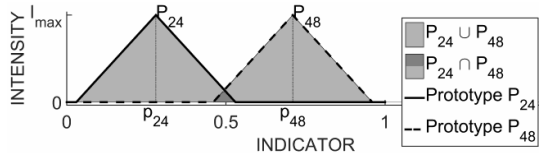

Figure 6: Similarity between two prototypes sampled at the 24$^{th}$ and 48$^{th}$ months of the time series, respectively.

The similarity between two shapes is $S(P_t, P_{t-24}) = P_t \cap P_{t-24}/P_t \cup P_{t-24}$. It is the ratio between the intersection and the union of the shapes. In Figure 6, $S(P_{48}, P_{24}) = 0.0097$.

*The Dissimilarity Subsystem*

This subsystem calculates the complement of the similarity between two prototypes generated at two instant of times, with a positive (negative) sign if the barycenter of the most recent track is larger (smaller) than the previous: $\Delta P = (1 - S(P_t, P_{t-2}) \cdot sgn(p_t - p_{t-2})$. In Figure 6, $\Delta P = 0.9903$. An unbiasing module is finally used, in order to provide one of three different classes as output: -1, +1, or 0, when the time series decreases, increases, or it is otherwise considered stable, respectively. In Figure 6, $\widetilde{\Delta P} = +1$, meaning that the indicator increases.

*The Adapting Subsystem*

The overall system uses 8 structural parameters (summarised in Table 1) to be appropriately adapted. Since different indicators in different Regions have different dynamics, manual adaptation is very time-consuming, human-intensive and error-prone (Ciaramella, 2010). In this section, we first describe the role of each parameter in the processing, and then we adopt a supervised optimization based on DE, an evolutionary technique for numerical optimization problems (Cimino, 2015).

Table 1: System Parameters.

| Module | Params | Human Expert | Range |
|---|---|---|---|
| marking | $\alpha_M\ \beta_M\ \varepsilon$ | (0.2; 0.8; 0.2) | (0, 1) |
| trailing | $\theta$ | (0.65) | (0, 1) |
| prototyping | $\alpha_P\ \beta_P$ | (0.15; 0.75) | (0, I$_{max}$) |
| dissimilarity | $\alpha_D\ \beta_D$ | (0.35; 0.65) | (0, 1) |

The *mark extension* ($\varepsilon$) controls the distance of interaction between marks. If it is close to 0, the marks cannot interact with each other, and there is no patterns reinforcement. If the mark extension is close to 1, all marks reinforce each other without distinction between patterns. The *mark evaporation* ($\theta$) affects the lifetime of a mark. Short-life marks evaporate to fast preventing aggregation and pattern reinforcement. Long-life marks cause an early saturation of the track, thus all tracks become similar to each other. Finally, α and β affect the system as previously described in the unbiasing module.

The adaptation uses the DE algorithm to optimize the parameters of the system with respect to the fitness computed using a training set. Let Z be the set of Regions in the dataset, $D_z(t_k)$ the input signal for Region $z$, $c_z(t_k) \in \{-1,0,1\}$ the output of the system. Let us consider $\{D_z(t_k)^*, C_z(t_k)^*\}$ the training set. We compute the average fitness among Regions as the Mean Squared Error (MSE) between the outputs of the system calculated for the training inputs and their corresponding training outputs:

$$f(Z) = \sum_{r=1}^{R} \sum_{k=4}^{Y} (c_z(t_k)^* - c_z(t_k))^2 / R \cdot Y.$$

In the DE algorithm, a solution is represented by a real *n*-dimensional vector, where *n* is the number of parameters to tune. The DE starts with a population of *N* candidate solutions, injected or randomly generated. In the literature different ranges of population are suggested (Mallipeddi, 2011). Population size spread can vary from a minimum of 2*n* to a maximum of 40*n*. A higher number increases the chance to find an optimal solution but it is more time consuming. To balance speed and reliability we use *N*=20. At each iteration and for each member (target) of the population, a mutant vector is created by mutation of selected members and then a trial vector is created by crossover of mutant and target. Finally, the best fitting among trial and target replaces the target.

Many strategies of the DE algorithm have been designed, by combining different structure and parameterization of mutation and crossover operators (Mezura, 2006 and Zaharie, 2007). We adopted the *DE/1/rand-to-best/bin* version, which places the perturbation at a location between a randomly chosen population member and the best population member. The differential weight $F \in [0,2]$ mediates the generation of the mutant vector. *F* is usually set in [0.4-1) (Mezura, 2006). There are different crossover methods in DE. Results show that a competitive approach can be based on binomial crossover (Zaharie, 2007). With binomial crossover, a component of the offspring is taken with probability *CR* from the mutant vector and with probability 1-*CR* from the target vector. A good value for *CR* is between 0.3 and 0.9 (Mallipeddi, 2011).

## 4 CASE STUDY AND RESULTS

The case study is based on a data set that contains the three annual indicators S, U, R, (described in Section 1) monitored for 15 years for 200 European Regions. The dataset contains 9000 samples. In order to reduce data to a canonical size, the following normalization is first applied: $x_n = (x_i - x_{min})/(x_{max} - x_{min})$. Since the original data samples are subject to significant sampling error, we also performed a granulation process (Cimino, 2014). More specifically, for each year we grouped regions via k-nearest neighbour algorithm. For each group we computed the annual mean μ and the standard deviation σ. We also determined that the resulting indicator samples are well-modelled by a normal distribution, using a graphical normality test. Finally, monthly samples have been derived considering normal distribution with mean and variance μ/12 and σ²/12, respectively.

To choose the best value of *CR* and *F*, we first performed trials with *CR* in [0.3, 0.6, 0.9] and *F* in [0.4, 0.6, 0.8]. For each experiment, 5 trials have been carried out, by using the 20% of the dataset as a training set, and the remaining 80% as a testing set. We also determined that the resulting MSE samples are well-modelled by a normal distribution, using a graphical normality test. Hence, we calculated the 95% confidence intervals. Table 2 shows the results in the form "mean ± confidence interval". The best performance has been with CR=0.6 and F=0.6. In general, we observed that fitness function gets stable after 15 generations only.

Table 2: 95% confidence interval of the MSE for the best setting of differential weight (F) and crossover rate (CR).

|    |     | F |  |  |
|----|-----|---|---|---|
|    |     | 0.4 | 0.6 | 0.8 |
| CR | 0.3 | 0.022 ± 0.0002 | 0.012 ± 0.00002 | 0.018 ± 0.0002 |
|    | 0.6 | 0.019 ± 0.0002 | **0.009± 0.0001** | 0.011 ± 0.00001 |
|    | 0.9 | 0.014 ± 0.00006 | 0.013 ± 0.00007 | 0.013 ± 0.00007 |

In order to assess the effectiveness of the approach, we adopted a 5-fold cross-validation. Indeed, each evaluation is also dependent on the data points, which end up in the training and test sets. For each trial, the training and test sets consist, respectively, of randomly extracted 20% and 80% of the original data. We carried out each trial 5 times. Table 3 summarizes, for indicator U, the results in terms of mean and standard deviation of the MSE for each trial. The low values of the MSE, for all trials and for both training and testing sets, highlight the effectiveness of the system in terms of both performance and generalization properties. We replicated the same experiments and achieved similar performances for the other indicators.

Finally, to highlight the great benefits of the adaptation subsystem, we also computed the MSE for the worst case of Table 2 (i.e., Trial 5), by using manual adaptation: this implied an MSE of 0.106, which is very higher than 0.022.

Table 3: MSE for each trial extracted via 5-fold cross-validation, averaged over 5 repetitions.

| Trial | MSE (mean ± std dev) | |
|---|---|---|
| | Training Set | Testing Set |
| 1 | 0.011 ± 0.010 | 0.018 ± 0.004 |
| 2 | 0.010 ± 0.010 | 0.020 ± 0.003 |
| 3 | 0.009 ± 0.006 | 0.020 ± 0.008 |
| 4 | 0.008 ± 0.008 | 0.020 ± 0.005 |
| 5 | 0.010 ± 0.007 | 0.022 ± 0.008 |

## 5 CONCLUSIONS

In this paper, we designed and developed a software system for assessing unfolding trends in innovation indicators. The core processing is based on stigmergy, a biologically inspired computational mechanism. Since the emergent character of stigmergy depends on biases and scale factors that can vary for different application contexts, an essential module is the parametric adaptation. For this purpose, we adopted the Differential Evolution. Experiments show the effectiveness of the approach and the relevant improvements with respect to a human parameterization.

More precisely the proposed system has been used to detect the trends of three different patent-based indicators within 35 technological domains, belonging to 268 European regions, in the period 1990-2012. The experimental results show that using 20% of the data set as training set to recognize trends ranging from -1 to 1, the system achieved an MSE of 0.02. Nevertheless, to ensure high-quality and robust design, the system should be cross-validated against other case studies and compared with existing approaches suitable for the same purpose. An important future development will be to adopt benchmark data and to carry out a comparative analysis of our approach with alternative techniques available in the literature.

## ACKNOWLEDGEMENTS

This work is supported by the University of Pisa, via the research project entitled "Stigmergic Footprint of Radical Innovations for Smart Specialisation in North-American and European Regions".